\documentclass[conference]{IEEEtran}
\IEEEoverridecommandlockouts
\usepackage{cite}
\usepackage{amsmath,amssymb,amsfonts}
\usepackage{algorithmic}
\usepackage{graphicx}
\usepackage{multirow}
\usepackage{textcomp}
\usepackage{xcolor}
\usepackage{caption}
\usepackage{subcaption}
\def\BibTeX{{\rm B\kern-.05em{\sc i\kern-.025em b}\kern-.08em
    T\kern-.1667em\lower.7ex\hbox{E}\kern-.125emX}}

\usepackage[pagebackref = true,
            breaklinks = true,
            letterpaper = true,
            colorlinks = true,
            bookmarks = false]{hyperref}

\newlength\savewidth\newcommand\shline{\noalign{\global\savewidth\arrayrulewidth\global\arrayrulewidth 1pt}\hline\noalign{\global\arrayrulewidth\savewidth}}

\def\mypar#1{\vspace{0mm}{\noindent\bf #1.}\hspace{1mm}}

\def\fig#1{Fig.~\ref{fig:#1}}
\def\tab#1{Tab.~\ref{tab:#1}}
\def\sect#1{Sec.~\ref{sec:#1}}

\def\eq#1{Eq.~(\ref{eq:#1})}

\usepackage{cuted}
\usepackage{capt-of}

\begin{document}

\title{\Huge 
Estimating 3D Uncertainty Field: \\Quantifying Uncertainty for Neural Radiance Fields
\thanks{Jianxiong Shen, Ruijie Ren and Francesc Moreno-Noguer are with Institut de Robòtica i Informàtica Industrial (CSIC-UPC), Spain. Adria Ruiz is with Seedtag, Spain. This work is supported partly by the Chinese Scholarship Council (CSC) under grant (201906120031). Contact: jshen@iri.upc.edu}
}

\author{\IEEEauthorblockN{Jianxiong Shen, Ruijie Ren, Adria Ruiz, Francesc Moreno-Noguer}}



\maketitle

\begin{strip}
    \vspace{-2cm}
    \centering
    \includegraphics[width=\textwidth]{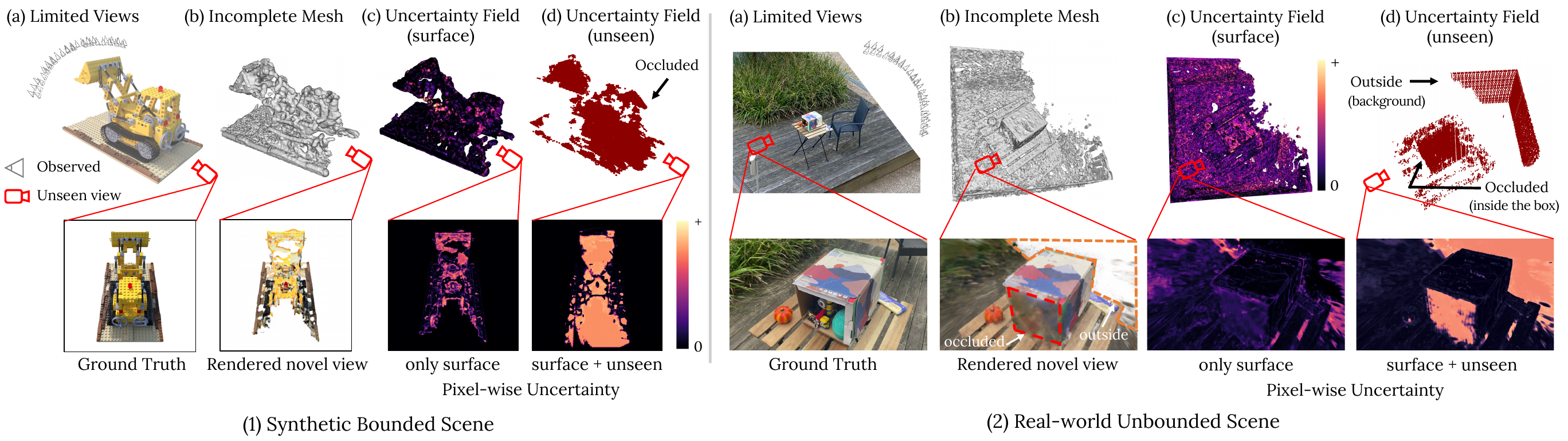}
    \captionof{figure}{Given observed images from limited view range (i.e. hemisphere), we propose to estimate a 3D \textit{Uncertainty Field} for quantifying the uncertainty in NeRF's predictions particularly on the occluded or outside scene content of (1) synthetic bounded scenes (i.e. back of truck) and (2) real-world unbounded scenes (i.e. inside the box), when modeling neural radiance field. This \textit{Uncertainty Field} can be naturally used for further robotic exploration and planning in the unknown scene regions. 
    \label{fig:teaser}}
\end{strip}

\begin{abstract}

Current methods based on Neural Radiance Fields (NeRF) significantly lack the capacity to quantify uncertainty in their predictions, particularly on the unseen space including the occluded and outside scene content. 
This limitation hinders their extensive applications in robotics, where the reliability of model predictions has to be considered for tasks such as robotic exploration and planning in unknown environments. To address this, we propose a novel approach to estimate a 3D \textit{Uncertainty Field} based on the learned incomplete scene geometry, which explicitly identifies these unseen regions. By considering the accumulated transmittance along each camera ray, our \textit{Uncertainty Field} infers 2D pixel-wise uncertainty, exhibiting high values for rays directly casting towards occluded or outside the scene content. To quantify the uncertainty on the learned surface, we model a stochastic radiance field. Our experiments demonstrate that our approach is the only one that can explicitly reason about high uncertainty both on 3D unseen regions and its involved 2D rendered pixels, compared with recent methods. Furthermore, we illustrate that our designed uncertainty field is ideally suited for real-world robotics tasks, such as next-best-view selection. 

\end{abstract}


\section{Introduction}

Neural Radiance Fields (NeRFs)~\cite{nerf} have emerged as a powerful tool for encoding intricate 3D scenes using only posed images, demonstrating remarkable capabilities and adaptability in different research fields \cite{d_nerf,pixelnerf,nerf_editing,DreamBooth} including robotics \cite{inerf,NeurAR,nerf_localization,nerf_grasp}. Such extensions have been explored to make the most of NeRF's effectiveness to solve various challenges in robotics such as SLAM\cite{nice_slam}, planning\cite{nerf_navigation} and manipulation\cite{nerf_supervision}. 

Despite their recent success in robotics, current NeRF-based methods still lack the ability to quantify the uncertainty associated with the model predictions, particularly in the areas of the scene where information is not provided during training. When robots implicitly use  NeRF-based approaches for exploring and planning in realistic environments, not being aware of predictions uncertainty can lead to catastrophic failures. 

Recently, some current methods have proposed to utilize ensemble-like methods like Deep Ensembles \cite{deep_ensemble} or MC-Dropout\cite{mc_dropout} to quantify NeRF uncertainty due to their model-agnostic property. Others \cite{snerf,cfnerf} have employed Variational Bayesian methods to model stochastic variants of NeRF which have been shown to provide better uncertainty measures. However, all these methods are inherently designed to estimate uncertainty for the areas of the scene that can be observed in the training images. As a consequence, for areas of the scene that are not provided during NeRF optimization, they tend to allocate low uncertainty values, that could lead to dangerous situations given that the robot is overconfident about unknown information about the scene. To overcome this, Niko et al. \cite{niko} compute an additional epistemic uncertainty along each ray concerning the unknown space. However, their ray-based method neglects the potential occluded regions and still predicts low uncertainty in non-observed areas. 

In this paper, we propose to estimate a 3D \textit{Uncertainty Field} along with the learned neural radiance fields. For the unseen regions including occluded and outside scene content from the observations, our uncertainty field presents high point-wise uncertainties and can also render high pixel-wise uncertainties for rays directly casting through these unknown regions (see \fig{teaser}). For estimating uncertainty on the learned geometry and appearance, we follow \cite{snerf} to model a stochastic radiance field based on the popular framework DVGO\cite{dvgo}. The evaluation results show that our approach achieves  state-of-the-art performance with reliable uncertainty on both synthetic bounded scenes and more challenging real-world unbounded scenes. The additional evaluation on the Next-Best-View planning task indicates that our estimated uncertainty field can be utilized for further exploration and planning in the unknown scene space.

\section{Related work}

\subsection{Neural Radiance Fields} 

NeRF \cite{nerf} implicitly encodes a 4D radiance field modelling a scene with a simple neural network from only 2D posed images. Due to its simplicity and impressive performance, it has gained significant popularity in various tasks \cite{nerf_diet,imap,Magic3D,MeshDiffusion,nerf_audio}. 
Recently, several extensions \cite{plenoxels,instantNGP,tensorf} have proposed to utilize different representations for faster convergence of NeRF. Among these methods, DVGO\cite{dvgo} directly optimizes the scene properties in voxel grids, achieving both efficient and accurate scene modelling. Additionally, its nature of modelling the scene in a normalized space, particularly for the unbounded scenes with complex backgrounds, provides a formal framework in our proposed approach for constructing the 3D uncertainty grid. 


\subsection{Uncertainty Quantification in Deep Learning} 

Uncertainty quantification is a crucial aspect in deep learning\cite{unc_review}, with applications in different fields including robotics\cite{unc_camera,unc_robotics}.
It provides a measure of how confident or uncertain a model is in its predictions.
In early methods, \textit{Bayesian Neural Networks} (BNN)~\cite{bnn_1} were employed to integrate uncertainty into the model by learning the marginal distributions of network weights.
Nonetheless, the training of BNNs can pose difficulties and demand significant computational resources, constraining their use in extensive deep neural network architectures.
Recently, several efficient strategies have been proposed to incorporate uncertainty estimation into deep neural networks like Deep Ensembles\cite{deep_ensemble} and MC-dropout\cite{mc_dropout}, due to their architecture-agnostic nature. 
However, their simple way of quantifying uncertainty by calculating the variance in the output space is not appropriate for identifying out-of-distribution data, particularly on 3D data\cite{niko,deep_ensemble}.

\subsection{Uncertainty Quantification in NeRF}

Recently, some works have explored different strategies to incorporate uncertainty estimation exclusively for NeRF. Among them, \cite{nerfw,activenerf,NeurAR} directly learned a uncertainty term in the RGB space by assuming a Gaussian distribution to predict pixel-wise uncertainties. Others \cite{UncertaintyGP,ActiveRMAP,ActiveIO} evaluate the volumetric uncertainty on the scene geometry by computing the information entropy from the predicted density along the ray. Instead, \cite{snerf,cfnerf} modeled a stochastic neural radiance field, allowing for quantifying uncertainty both on 3D scene geometry and 2D rendered images. However, without the knowledge in the unknown scene regions, all these NeRF-based methods are unable to explicitly recognize them and allocate high uncertainties there. To address it, Niko et al.\cite{niko} consider an additional uncertainty measure in the unknown space according to the ray termination probability on the learned geometry. However, they neglect the existing occlusion in most scenes, which can be crucial in practical cases. To the best of our knowledge, our approach is the first one that can not only quantify the uncertainty on the learned geometry and appearance, but also allocate high uncertainties on unseen regions, including outside and occluded points. 

\section{Uncertainty Quantification}

\subsection{Background}

NeRF \cite{nerf} based methods represent a scene as, 
\begin{align}
    (\sigma, \mathbf{c}) = f_{\theta}(\mathbf{x},\mathbf{d}) \,,
\end{align}
where two scene attributes, density $\sigma$ and radiance $\mathbf{c}$ for each input 3D coordinate $\mathbf{x}$ and view direction $\mathbf{d}$ are modeled by a 5D function: $\mathbb{R}^5 \mapsto \mathbb{R}^4$. The volume rendering technique is then used to render the pixel color: 
\begin{align}
    \label{eq:volume_render}
    \hat{C}(r) = \int_{t_n}^{t_f}T(t)(1-e^{-\sigma_t\delta}) c_t dt, \,\,\, T(t)=\prod_{i=t_n}^{t}e^{-\sigma_i\delta} \,,
\end{align} 
where $T(t)$ denotes the accumulated transmittance from $t_n$ to $t_f$ representing the probability of not hitting any scene objects.

DVGO \cite{dvgo} explores a more efficient hybrid representation for fast convergence in NeRF models. With the scene bounded in a normalized space (i.e. a cube), they directly utilize voxel grids $\mathbf{V}_{den}$ and $\mathbf{V}_{rad}$ with resolution $N_x \times N_y \times N_z$ to represent $\sigma, \mathbf{c}$ respectively. The specific value for arbitrary $(\mathbf{x}, \mathbf{d})$ can be obtained by trilinear interpolation as, 
\begin{align}
    \sigma &= \text{interp}(\mathbf{x},\mathbf{V}_{den}) \nonumber \\
    \mathbf{c} &= \text{Light-weight-MLP}(\text{interp}(\mathbf{x},\mathbf{V}_{rad}),\mathbf{d}) \,.
\end{align}
Please refer to their paper \cite{dvgo} for more details. 
Explicitly storing the density value in a voxel grid is not only efficient in optimization, but also makes it tractable to intuitively distinguish the different regions (i.e. occluded) by voxels.

\subsection{Predictive Uncertainty for In-Distribution points}
\label{sec:predictive_uncertainty}

Training NeRF models involves ray casting and point sampling through the scene. For clear illustration, we first define different types of points and rays, which contribute differently in reconstructing a scene by NeRF-based methods, as shown in \fig{definition_comparison}. Those points, which can be seen in the observed reference images, are known as In-distribution points and can be split into two types: \textbf{Empty points} (E) falling inside the unoccupied voxels of low or zero density values and \textbf{Surface points} (S) falling inside the informative voxels representing the scene geometry. The scene attributes $(\sigma,\mathbf{c})$ for these two types of points are updated in the optimization process, and thereby its corresponding uncertainties (often called aleatoric uncertainty) can be learned by the model parameters. To estimate the uncertainty for points \textbf{E} and \textbf{S}, we model a stochastic radiance field based on DVGO\cite{dvgo} following \cite{snerf,cfnerf}. Concretely, we construct a density voxel grid $\mathbf{V}_{den} \in \mathbb{R}^{N_x \times N_y \times N_z \times 2}$ where each vertex stores parameters $(\mu_{\sigma}, \beta_{\sigma}^2)$ of a Gaussian distribution. The estimated radiance color is also a Gaussian $(\mu_{c}, \beta_{c}^2)$. Their specific values for arbitrary input pairs $(\mathbf{x}, \mathbf{d})$ can be obtained as, 
\begin{align}
    \label{eq:interpolation}
    (\mu_{\sigma}, \beta_{\sigma}^2) &= \text{interp}(\mathbf{V}_{den}, \mathbf{x}) \, , \nonumber \\
    ({\mu}_{\mathbf c}, \beta_{\mathbf c}^2) &= \text{Light-weight-MLP}(\text{interp}(\mathbf{V}_{rad}, \mathbf{x}), {\mathbf d}) \,.
\end{align}

Reparameterization trick can be then used to draw sampled density/color from the probabilistic radiance field, $\sigma / \mathbf{c} = \mu+\epsilon\beta$ where $\epsilon \sim \mathcal{N}(0,1)$. In this manner, multiple (i.e. K) possible $\alpha$-distance trajectories for each ray can be obtained, followed by \eq{volume_render} to generate multiple pixel color estimates $\hat{C}_{1:K}$. Note that $\epsilon$ is shared by all the points on each trajectory. Then, the variance of the multiple estimates can be treated as the associated uncertainty $\mathbf{U}_C(r)=\text{Var}(\hat{C}_{1:K})$. 

As for optimization, we follow \cite{snerf,cfnerf} to minimize the negative log-likelihood $-\log p(\hat{C}_{k}=C^{*}|\mu_{\sigma},\beta_{\sigma},\mu_{\mathbf{c}},\beta_{\mathbf{c}})$ given ground-truth pixel RGB values $C^{*}$. We then utilize multivariate kernel density estimation\cite{kde} to approximate the probability density and formulate the pixel-wise loss as,
\begin{align}
    \mathcal{L} = -\log \frac{1}{K}\sum_{k=1}^{K}(2\pi)^{-\frac{3}{2}}|\mathbf{H}|^{-\frac{1}{2}}e^{-\frac{1}{2}\mathbf{C}^\text{T}_{k}\mathbf{H}^{-1}\mathbf{C}_{k}} + \frac{\lambda}{K}\sum_{k=1}^K \sigma_{k}\;,
\end{align}
where $\mathbf{C}_{k}=\hat{C}_{k}-C^{*}$ and the bandwidth matrix $\mathbf{H}$ can be empirically chosen as $0.98\frac{Var(\hat{C}_{1:K})}{K^{1/7}}$. The second term is often adopted as a regularizer\cite{nerfw}.

\subsection{Uncertainty Field for Out-of-Distribution points}

\begin{figure}[t]
    \centering
    \begin{subfigure}{0.4\textwidth}
        \centering
        \includegraphics[width=\linewidth]{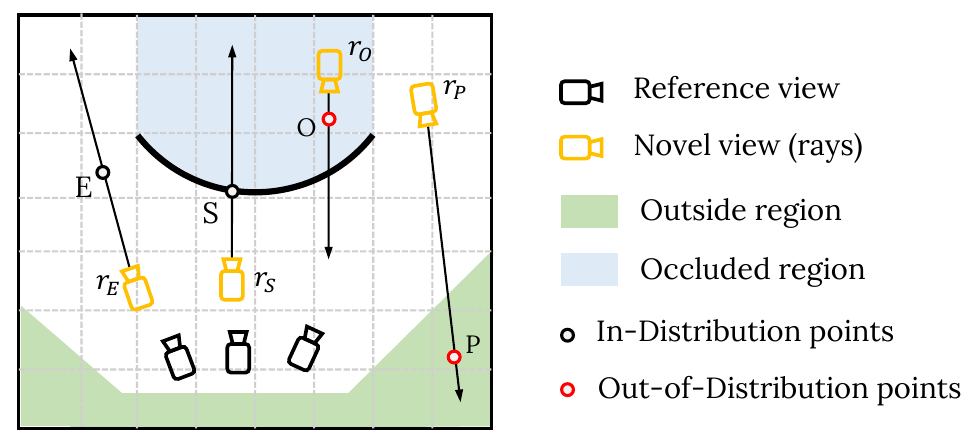}
    \end{subfigure}%
    \vspace{3mm}
    \begin{subfigure}{\linewidth} 
        \centering
        \small
        \setlength{\tabcolsep}{1.2pt} 
        \renewcommand{\arraystretch}{1.25} 
        \begin{tabular}{lccccc}
        \shline
        \multirow{2}{*}{\textbf{Methods}} & \multicolumn{2}{c}{\textbf{In-Distribution}} &  & \multicolumn{2}{c}{\textbf{Out-of-Distribution}} \\ \cline{2-3} \cline{5-6} 
         & Empty(E) & Surface(S) &  & Outside(P) & Occluded(O) \\ \hline
        Ensemble-like\cite{deep_ensemble} & $\checkmark$ & $\checkmark$ &  &  &  \\
        MC-Dropout\cite{mc_dropout} & $\checkmark$ & $\checkmark$ &  &  &  \\
        S-NeRF\cite{snerf} & $\checkmark$ & $\checkmark$ &  &  &  \\
        CF-NeRF\cite{cfnerf} & $\checkmark$ & $\checkmark$ &  &  &  \\
        Niko et al.\cite{niko} & $\checkmark$ & $\checkmark$ &  & $\checkmark$ &  \\
        Ours & $\checkmark$ & $\checkmark$ &  & $\checkmark$ & $\checkmark$ \\ \shline
        \end{tabular}%
    \end{subfigure}
    \caption{(Top) We show the illustration of different point types when exploring a scene with unseen regions, and (Bottom) whether current methods are able to estimate uncertainty on these points.}
    \label{fig:definition_comparison}
\end{figure}

Apart from the \textit{in-distribution} points that can be seen from the observed views, there exist also \textit{out-of-distribution} points that cannot be seen from the observed views, as shown in \fig{definition_comparison}(Top). 
The points in these unseen regions can be further split into two types: \textbf{Outside points} (P) falling outside the observed views, and \textbf{Occluded points} (O) caused by occlusion of the scene objects. We compare the ability of current methods to estimate uncertainty on all four points in \fig{definition_comparison}(Bottom). 
NeRF-based models including ours will not update the density values for these \textit{out-of-distribution} points in these unseen regions and thereby always predict low uncertainties despite of high predictive errors.
Niko et al. \cite{niko} consider the outside points, but still cannot allocate high uncertainty for occluded points (see \fig{qualitative_results_2D}). 

Theoretically, these regions that are not seen during training should all be allocated with high uncertainty. This problem could then be transformed to distinguish the \textit{out-of-distribution} points from these determined empty and surface points. For this purpose, we introduce a model-free \textit{Uncertainty Field}, implemented by an additional uncertainty voxel grid denoted as $\mathbf{V}_{H}$ with the same resolution with $\mathbf{V}_{den}$. Before training, we initialize $\mathbf{V}_{H}=1$, indicating that the scene is full of high uncertainties everywhere. During training, the empty points seen in the training images will be considered unoccupied, resulting in a constant accumulated transmittance $T(\textbf{E}) = 1$. When the ray travels to the scene object surface, the intersected points there will be considered occupied with allocated increased high density values, resulting in sharply decayed $T(\textbf{S})$ ranging from 1 to $\tau$ (i.e. $\tau=0.1$). Behind the object surface, the points with $T(\textbf{O}) < \tau$ can be considered occluded and not seen from the rays. We update the  uncertainty field $\mathbf{V}_H$ as,
\begin{align}
    \label{eq:update_V_H}
    \mathbf{V}_H(\mathbf{x}_t) = 
    \begin{cases}
      \text{0} & \text{if } \, T(t)>\tau \,,\\
      \text{1} & \text{if } \, T(t)<\tau \,.\\
    \end{cases}
\end{align}
The remaining unsampled regions in the 3D bounding box during training are treated as outside (P), where $\mathbf{V}_H$ will not be updated and remain to be of high uncertainty. The above updating operation involves no parameters to learn and thereby could be fully processed offline after training.
Finally, the complete 3D point-wise uncertainty could be obtained as,
\begin{align}
    \mathbf{U}(\mathbf{x}) = \text{interp}(\mathbf{V}_H, \mathbf{x}) + \text{Var}(\text{softplus}(\sigma_{\mathbf{x}}))
\end{align}
where the second term computes the  uncertainty on the surface points brought by our modelled density distribution. 

\subsection{Inferring 2D Pixel-wise Uncertainty}

Given the updated $\mathbf{V}_H$, it is also convenient to infer the accumulated uncertainty along each camera ray for each rendered 2D pixel. Whether the high uncertainties in the unseen regions emerge in the 2D uncertainties associated with the pixel color depends on the specific ray direction. Along with the defined four types of points in the scene in \fig{definition_comparison}, we also define four corresponding rays $\mathbf{r}_E, \mathbf{r}_S, \mathbf{r}_O, \mathbf{r}_P$. The computed 2D accumulated uncertainties from $\mathbf{V}_H$ should satisfy the following requirements: 

1) From the view direction of the first two rays, $\mathbf{r}_E$ and $\mathbf{r}_S$, both the occluded and outside points are not visible. As a result, the high uncertainty brought by these regions should not be considered, in short, $\mathbf{U}_H(r_E)=0$ and $\mathbf{U}_H(r_S)=0$. 

2) On the contrary, for $\mathbf{r}_O$ and $\mathbf{r}_P$, either the occluded or the outside points are visible. Consequently, their accumulated uncertainties $\mathbf{U}_H(r_O)$ and $\mathbf{U}_H(r_P)$ should be high. 

\begin{figure}
    \centering
    \includegraphics[width=\linewidth]{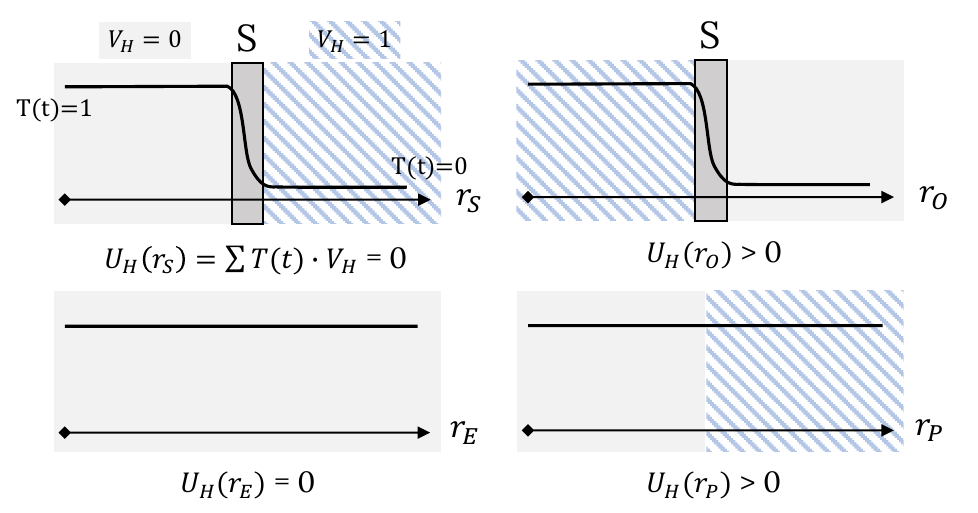}
    \caption{The inferred pixel-wise uncertainty from our uncertainty field exhibits high value for rays that directly cast towards the occluded $r_O$ or outside scene content $r_P$. $T(t)$ represents the accumulated transmittance and will quickly decay to zero when hitting the surface (S).}
    \label{fig:infer_2D_uncertainty}
\end{figure}

To achieve the above two requirements, we introduce a weighted uncertainty accumulated along each camera ray to take the ray termination into consideration. Formally, we compute it as, 
\begin{align}
    \mathbf{U}_H(r) = \sum_{t=t_n}^{t_f} T(t)\cdot\text{interp}(\mathbf{V}_H, \mathbf{x}_t) \,,
\end{align}
where $T(t)$ is the accumulated transmittance introduced in \eq{volume_render}. To avoid too big values, we normalize it using $1-\text{exp}(-\mathbf{U}_H(r))$ as the final result. 
The specific analysis on inferring $\mathbf{U}_H(r)$ for all the four types of rays is demonstrated in \fig{infer_2D_uncertainty}. 
Finally, the complete 2D pixel-wise uncertainty along a camera ray is equivalent to:
\begin{align}
    \label{eq:uncertrainty_terms}
    \mathbf{U}(r) = \mathbf{U}_C(r) + \mathbf{U}_H(r) \,,
\end{align}
where $\mathbf{U}_C(r)$ is the predictive uncertainty from our stochastic radiance fields in \sect{predictive_uncertainty}.

\section{Application: Next Best View Planning}
\label{sec:application}

Our designed uncertainty field is ideally suited for robotic exploration and planning in unknown environments. A critical task in robotics, known as Next-Best-View (NBV) selection, involves identifying the most informative views to effectively represent unseen scene content.

Starting from a limited set of initial observations of a scene, there always remain unseen or occluded regions within the unknown scene space that require exploration. Given resource constraints, the task is to identify the next most informative views that contain more features to aid in scene reconstruction. Our estimated uncertainty field is naturally suited for this task.
In the first step, we train our model using the initial set of observed images. Based on the learned geometry, we estimate an uncertainty field where unseen regions maintain a high level of uncertainty, as defined by \eq{update_V_H}. In the second step, we evaluate all candidate views for their 2D image-wise uncertainty, $\mathbf{U}_H(\mathcal{I})$, which is calculated as the sum of pixel-wise uncertainties within each image:
\begin{align}
    \mathbf{U}_H(\mathcal{I}) &= \sum_{r \in \mathcal{I}} \mathbf{U}_H(r) \,,
    \\
    \mathcal{I}_{\text{NBV}} &= \arg \max_{\mathcal{I}} \mathbf{U}_H(\mathcal{I}) \,,
\end{align}
where the selected next best view, $\mathcal{I}_{\text{NBV}}$, is the one with the maximum uncertainty value. We then update the uncertainty grid, $V_H$, using the rays cast and points sampled through the selected $\mathcal{I}_{\text{NBV}}$, without the need for additional training.
With the updated uncertainty grid, we repeat the second step until we have identified N target views to add to the training set. We then retrain the model and iterate this process until we reach the resource budget.

\begin{figure*}[t]
    \begin{subfigure}{\textwidth}
    \centering
    \includegraphics[width=\linewidth]{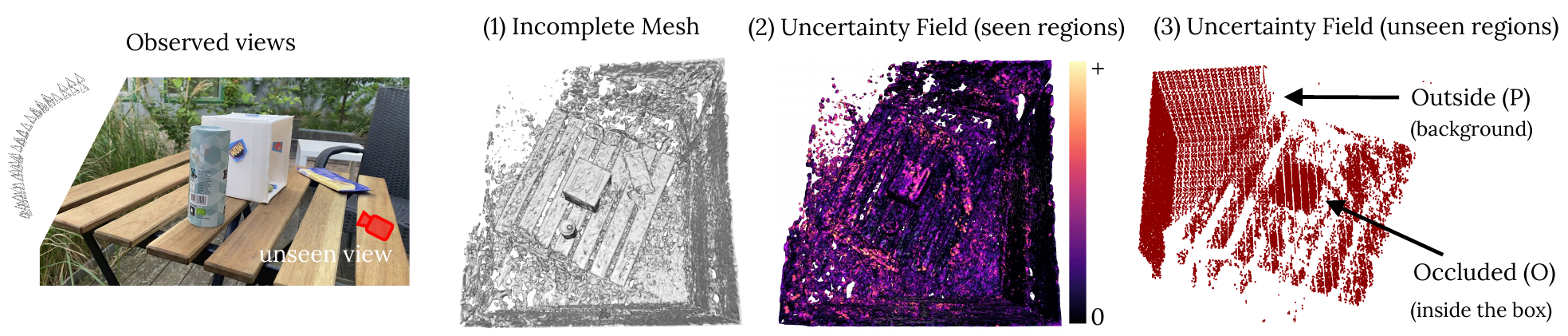}
    \caption{Example of our estimated 3D point-wise uncertainty. Given observed images only from the left hemisphere, our approach can not only infer the (2) predictive uncertainty for the learned incomplete scene geometry, but also allocate high uncertainties on (3) unseen regions, including outside (i.e. the background) and occluded points (i.e. inside the box) shown in red color.}
    \label{fig:qualitative_results_3D}
    \end{subfigure}
    \begin{subfigure}{\textwidth}
    \centering
    \includegraphics[width=\linewidth]{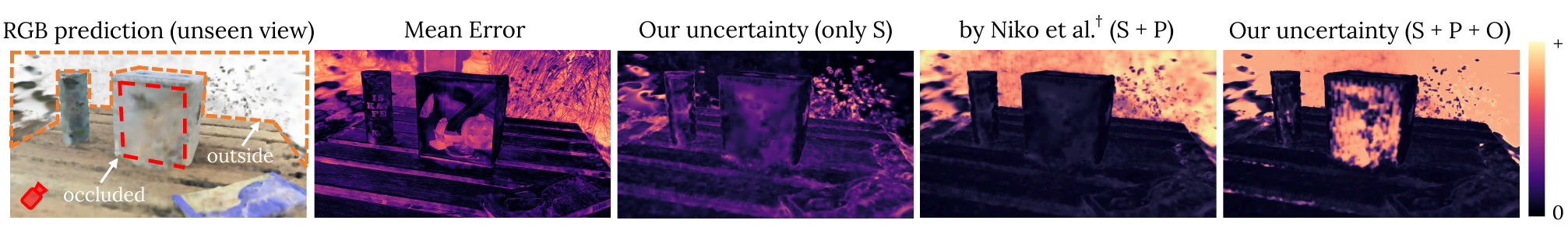}
    \caption{Example of our inferred 2D pixel-wise uncertainty. For the rendered RGB prediction from an unseen view with occlusion and outside scene content, our method is the only one that can correctly estimate uncertainty on all Surface (S), Outside (P) and Occluded (O) points.}
    \label{fig:qualitative_results_2D}
    \end{subfigure}
    \caption{Qualitative results of our estimated (a) point-wise and (b) pixel-wise uncertainty.
    Without the knowledge in the unseen scene regions, the model tends to allocate zero density there. Rendering through these regions produces notably incorrect pixel colors exhibiting purely white or transparency, and correspondingly high uncertainty estimates (yellower) from our approach. }
    \label{fig:qualitative_results}
\end{figure*}

\begin{table*}[h]
\centering
\setlength{\tabcolsep}{6pt} 
\renewcommand{\arraystretch}{1.25} 
\caption{Evaluation results on the pixel-wise uncertainty estimates. Our approach achieves notably more reliable uncertainty measures than other baselines across both the synthetic bounded and realistic unbounded scenes. We report the Area Under the Sparsification Error curve (AUSE) as the evaluation metric. Best results in Bold.}
\resizebox{0.95\textwidth}{!}{%
\begin{tabular}{clccccccccccccc} 
\shline
 & \multirow{2}{*}{Methods} &  & \multicolumn{6}{c}{NeRF-Synthetic (bounded)} &  & \multicolumn{5}{c}{Real-Occlusion (unbounded)} \\
 & &  & Chair & Lego & Drums & Hotdog & Materials & Average &  & Indoor1 & Indoor2 & Outdoor1 & Outdoor2 & Average \\ 
 \hline 
\multirow{5}{*}{AUSE $\downarrow$} & D.E.\cite{deep_ensemble} &  & 0.15 & 0.22 & \textbf{0.16} & 0.23 & 0.10 & 0.17 &  & 0.48 & 0.56 & 0.73 & 0.77 & 0.63 \\
& S-NeRF\cite{snerf} &  & 0.17 & 0.31 & 0.22 & 0.30 & 0.12 & 0.22 &  & 0.47 & 0.53 & 0.78 & 0.81 & 0.67 \\ 
& CF-NeRF \cite{cfnerf} &  & 0.16 & 0.29 & 0.24 & 0.31 & 0.12 & 0.22 &  & 0.29 & 0.33 & 0.49 & 0.52 & 0.41 \\
& Niko et al.$\textsuperscript{\dag}$\cite{niko} &  & 1.33 & 1.09 & 1.29 & 1.12 & 1.04 & 1.17 &  & 0.32 & 0.27 & 0.31 & 0.33 & 0.31 \\
& Ours $(U_C + U_H)$ &  & \textbf{0.14} & \textbf{0.13} & 0.17 & \textbf{0.21} & \textbf{0.09} & \textbf{0.15} &  & \textbf{0.26} & \textbf{0.22} & \textbf{0.19} & \textbf{0.18} & \textbf{0.21} \\ 
\hline 
\multirow{2}{*}{Ablation study} & Ours $(U_C)$ &  & 0.18 & 0.30 & 0.23 & 0.32 & 0.09 & 0.22 &  & 0.30 & 0.35 & 0.51 & 0.48 & 0.41 \\
& Ours $(U_H)$ &  & 0.24 & 0.21 & 0.36 & 0.22 & 0.44 & 0.29 &  & 0.64 & 0.61 & 0.53 & 0.53 & 0.58 \\
\shline
\end{tabular}%
}
\label{tab:results}
\end{table*}

\section{Experimental Evaluation}

\subsection{Evaluation on Uncertainty Quantification}

\mypar{Datasets} Our main goal is to estimate the quality of the provided uncertainty measures concerning the unseen regions, particularly in the occluded regions of the scene. For this reason, we collect a real-world unbounded dataset, including two indoor and two outdoor scenes with an image resolution of 950x540. All the scenes are captured in an inward-facing manner towards the objects, following the protocol in \cite{mipnerf360}. In each scene, all sides of the object are closed except for one opening, causing internal occlusion. We denote this dataset as \textbf{Real-Occlusion}. To conduct evaluation, we first sample testing views uniformly around the scene. Among the remaining views, we randomly select training views in the hemisphere on the opposite of the opening side, guaranteeing that no information inside is observed(see \fig{qualitative_results_3D}). Additionally, to evaluate our method on bounded scenes, we also select five scenes with obvious occlusion from \textbf{NeRF-Synthetic} dataset\cite{nerf}. For training, we randomly select training views from a narrow range (i.e. from a quarter) and use the remaining views for testing. 

\mypar{Baselines, Metrics} We first consider a general model-agnostic method, Deep Ensembles(D.E.), and two current popular methods designed to exclusively estimate uncertainty in NeRF-based models, S-NeRF and CF-NeRF. Additionally, we compare against another recent method \cite{niko}, exclusively designed for estimating uncertainty on unseen regions. To implement it and D.E., we follow the same setting in \cite{niko} and train five models in the ensemble based on DVGO\cite{dvgo}.
To evaluate the correlation between the predicted error and estimated uncertainty, we report the Area Under the Sparsification Error (AUSE) curve\cite{ause}. A lower AUSE value indicates more reliable uncertainty estimates. 

\mypar{Implementation} To implement our method, we mainly inherit the dense voxel grid structure and hyperparameters from \cite{dvgo}. 
We set 0.01 as the initialized learning rate of the additional density variance. The weight of regularizer $\lambda$ is set to be 0.001 for all scenes unless specified. 
We train our model on each scene for 10000 iterations on a single Nvidia RTX 2080Ti, which takes around 10 and 15 mins for NeRF-Synthetic and our Real-Occlusion dataset respectively. 

\mypar{Results} 
In \tab{results}, we show the quantitative results of different methods both on bounded and unbounded scenes. Most previous methods including D.E., S-NeRF and CF-NeRF are designed only to estimate uncertainty for the known geometry seen from observed regions. On the unknown space, they tend to allocate constantly zero density, producing high-error predictions with low uncertainty estimates there. As a result, their AUSE values are significantly higher, particularly on challenging realistic unbounded scenes. Although Niko et al. consider the uncertainty in outside scene regions, their proposed approach still neglects the occluded points, resulting in inferior performance than ours on \textit{Real-Occlusion} unbounded scenes. As a consequence, their method tends to allocate high uncertainty also for white background pixels involved in bounded scenes, leading to worst results. In contrast, our approach achieves the lowest AUSE values of 0.15 and 0.21 averagely on synthetic and realistic scenes, indicating that our learned 3D uncertainty field can produce highly consistent pixel-wise uncertainty.
A qualitative analysis of our uncertainty estimation results is demonstrated in \fig{qualitative_results}. 

\subsection{Ablation Studies} 

\mypar{Effect of individual uncertainty term} 
As described in \eq{uncertrainty_terms}, our inferred 2D pixel-wise uncertainty is composed of two terms: the variance in the RGB space $U_C$ estimating the uncertainty on the learned geometry, and the accumulated post-uncertainty $U_H$ accounting for the unseen scene regions. As can be seen at the bottom of \tab{results}, neither of them can be used alone to explain reliable uncertainty for the rendered novel views. 

\begin{figure}
    \centering
    \includegraphics[width=\linewidth]{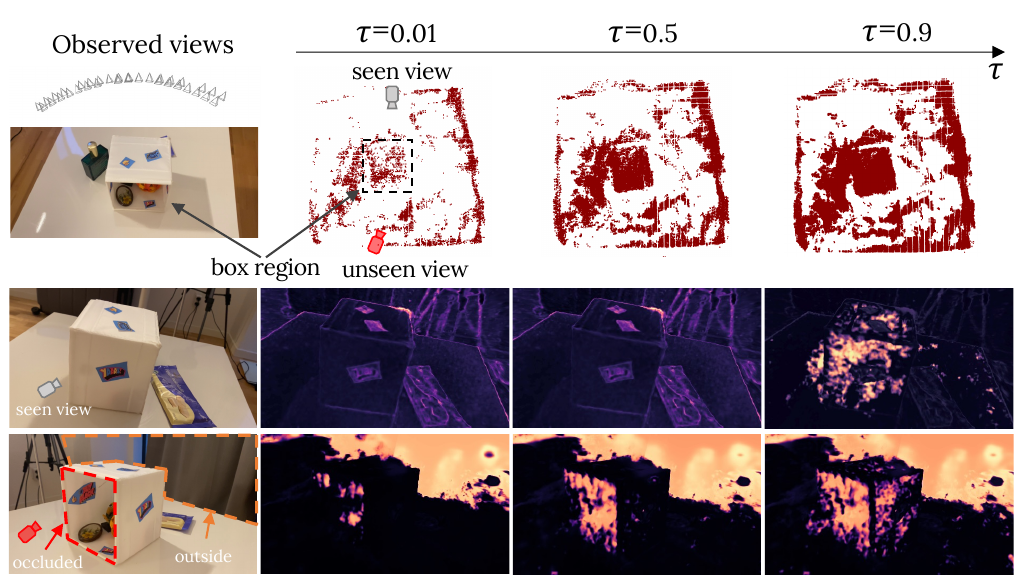}
    \caption{Ablation study on the threshold $\tau$ used to distinguish the occluded points from the seen space. Too small $\tau$ causes low inferred pixel-wise uncertainties for rays casting through the unseen occluded scene regions, while too large $\tau$ will destroy the reasonable uncertainty from the seen view.}
    \label{fig:ablation_thres}
\end{figure}

\mypar{Effect of the threshold $\tau$} 
As described in \eq{update_V_H}, $\tau$ is used as a threshold to distinguish the occluded points from the seen space. Theoretically, higher $\tau$ means that more points beyond the seen surface will be treated as occluded, exhibiting high values in the uncertainty field $\mathbf{V}_H$. As shown in \fig{ablation_thres}, when we use a relatively smaller $\tau$ of 0.01, the estimated uncertainty field exhibits fewer points of high uncertainty in the occluded space - inside the box. As a result, the rendered 2D pixel-wise uncertainties for the rays that cast directly towards the occluded regions present unreliable low values. 
On the contrary, when we set a large value of 0.9 for $\tau$, nearly all points inside the box are exhibiting high uncertainties in the estimated uncertainty field. Although reasonable on these occluded pixels after rendering, its rendered 2D pixel-wise uncertainties on the seen surface exhibit exceptional high values, which should be low as expected. In sum, choosing a proper value for $\tau$ is crucial for inferring reliable pixel-wise uncertainties on both seen and unseen views.

\begin{figure}
    \centering
    \includegraphics[width=\linewidth]{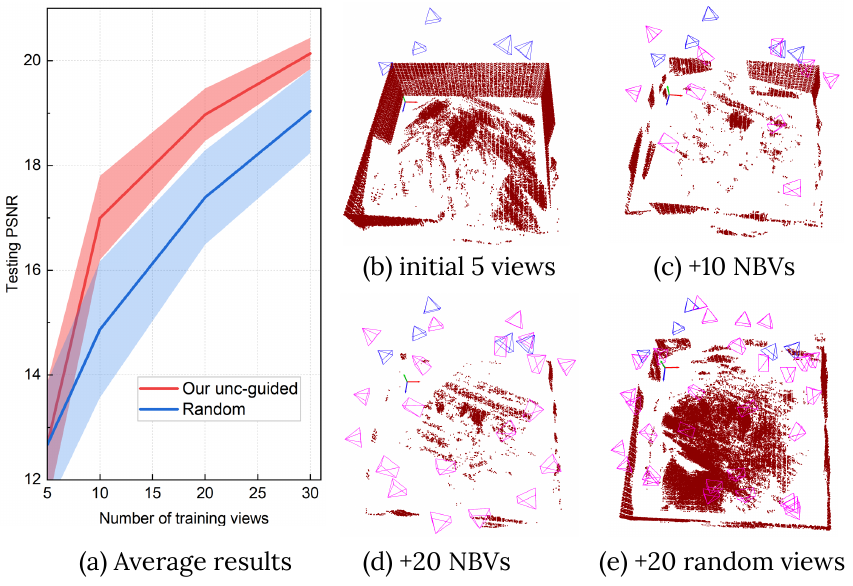}
    \caption{(a) Starting from 5 views in a hemisphere, we progressively add new views, selected either randomly or based on our uncertainty measures. Results are averaged over the Real-Occlusion dataset. (b) The estimated uncertainty field given initialized random 5 views for the \textit{Outdoor1} scene. The next-best-views (NBVs) selected by our uncertainty-guided strategy effectively target the unknown scene space, as evidenced by the significant reduction in high-uncertainty regions in the uncertainty field after adding (c) 10 and (d) 20 more NBVs. In contrast, (e) shows large unobserved regions despite the addition of 20 random views.}
    \label{fig:next_best_views}
\end{figure}

\subsection{Evaluation on Next-Best-View Selection}

We now evaluate our uncertainty-guided strategy utilizing our estimated uncertainty field in a real-world application: Next-Best-View (NBV) selection. Considering the practical environments are unbounded with complex backgrounds and full of occlusions, we mainly evaluate the \textit{Real-Occlusion} dataset, which is challenging for most current methods. We first split around 12\% of testing views uniformly sampled around the scene. From the hemisphere on the opposite of the opening side in each scene, we randomly select initialized training images with a number of 5. All the remaining views are treated as candidate views for further selection. We follow the reconstruction process described in \sect{application} with $N=10$. After each iteration, we report PSNR to evaluate the quality of rendered novel images. For a baseline, we mainly compare our uncertainty-guided approach against a general selection strategy: \textit{Randomly} selecting images from the candidate views. 

\mypar{Results} 
\fig{next_best_views}(a) shows the results during the next-best views searching process, averagely on \textit{Real-Occlusion} dataset. It clearly shows that the selected views by our estimated uncertainty are significantly more informative than simple random selection. From the qualitative analysis in \fig{next_best_views}(b-e), our selected NBVs effectively target at those unknown scene regions and thereby provide more information for better reconstructing the scene. 

\section{Conclusion and future work} 

We have proposed to estimate a 3D \textit{Uncertainty Field} to explicitly allocate high uncertainties for the occluded and outside scene content in reconstructing a real-world scene. Beside, we model a stochastic radiance field of the scene based on the observed images to estimate the predictive uncertainties on the learned scene geometry and appearance. The further rendering through our \textit{Uncertainty Field} clearly produces reliable 2D pixel-wise uncertainty estimates on both observed and unseen regions. Furthermore, our evaluation on a real-world robotics task show the effectiveness of our \textit{Uncertainty Field} in selecting the most informative next views when exploring the unknown scene space. 

As one of the major advantages, our introduced \textit{Uncertainty Field} could be fully estimated offline after training, allowing to be conveniently integrated in any trained NeRF-based models for recognizing and exploring the unobserved scene regions. 

Limited by the current scene parameterization \cite{dvgo_v2,nerf++}, the current structure of our \textit{Uncertainty Field} is mainly designed for object-centric scenes. In the future, more general structure is expected to be developed for robotic exploration in more general scenes with free camera trajectory. 

 


\clearpage

\ifCLASSOPTIONcaptionsoff
  \newpage
\fi

{\small
\bibliographystyle{IEEEtran}
\bibliography{references}

\begin{thebibliography}{10}
\providecommand{\url}[1]{#1}
\csname url@samestyle\endcsname
\providecommand{\newblock}{\relax}
\providecommand{\bibinfo}[2]{#2}
\providecommand{\BIBentrySTDinterwordspacing}{\spaceskip=0pt\relax}
\providecommand{\BIBentryALTinterwordstretchfactor}{4}
\providecommand{\BIBentryALTinterwordspacing}{\spaceskip=\fontdimen2\font plus
\BIBentryALTinterwordstretchfactor\fontdimen3\font minus
  \fontdimen4\font\relax}
\providecommand{\BIBforeignlanguage}[2]{{%
\expandafter\ifx\csname l@#1\endcsname\relax
\typeout{** WARNING: IEEEtran.bst: No hyphenation pattern has been}%
\typeout{** loaded for the language `#1'. Using the pattern for}%
\typeout{** the default language instead.}%
\else
\language=\csname l@#1\endcsname
\fi
#2}}
\providecommand{\BIBdecl}{\relax}
\BIBdecl

\bibitem{nerf}
B.~Mildenhall, P.~P. Srinivasan, M.~Tancik, J.~T. Barron, R.~Ramamoorthi, and
  R.~Ng, ``Nerf: Representing scenes as neural radiance fields for view
  synthesis,'' in \emph{ECCV}, 2020.

\bibitem{d_nerf}
A.~Pumarola, E.~Corona, G.~Pons-Moll, and F.~Moreno-Noguer, ``{D-NeRF}: Neural
  radiance fields for dynamic scenes,'' \emph{2023 IEEE/CVF Conference on
  Computer Vision and Pattern Recognition (CVPR)}, 2021.

\bibitem{pixelnerf}
A.~Yu, V.~Ye, M.~Tancik, and A.~Kanazawa, ``pixelnerf: Neural radiance fields
  from one or few images,'' \emph{2021 IEEE/CVF Conference on Computer Vision
  and Pattern Recognition (CVPR)}, pp. 4576--4585, 2020.

\bibitem{nerf_editing}
Y.-J. Yuan, Y.~tian Sun, Y.-K. Lai, Y.~Ma, R.~Jia, and L.~Gao, ``Nerf-editing:
  Geometry editing of neural radiance fields,'' \emph{2022 IEEE/CVF Conference
  on Computer Vision and Pattern Recognition (CVPR)}, pp. 18\,332--18\,343,
  2022.

\bibitem{DreamBooth}
N.~Ruiz, Y.~Li, V.~Jampani, Y.~Pritch, M.~Rubinstein, and K.~Aberman,
  ``Dreambooth: Fine tuning text-to-image diffusion models for subject-driven
  generation,'' \emph{2023 IEEE/CVF Conference on Computer Vision and Pattern
  Recognition (CVPR)}, pp. 22\,500--22\,510, 2022.

\bibitem{inerf}
Y.-C. Lin, P.~R. Florence, J.~T. Barron, A.~Rodriguez, P.~Isola, and T.-Y. Lin,
  ``inerf: Inverting neural radiance fields for pose estimation,'' \emph{2021
  IEEE/RSJ International Conference on Intelligent Robots and Systems (IROS)},
  pp. 1323--1330, 2020.

\bibitem{NeurAR}
Y.~Ran, J.~Zeng, S.~He, L.~Li, Y.~Chen, G.~H. Lee, J.~Chen, and Q.~Ye,
  ``Neurar: Neural uncertainty for autonomous 3d reconstruction,''
  \emph{ArXiv}, vol. abs/2207.10985, 2022.

\bibitem{nerf_localization}
D.~Maggio, M.~Abate, J.~Shi, C.~Mario, and L.~Carlone, ``Loc-nerf: Monte carlo
  localization using neural radiance fields,'' \emph{2023 IEEE International
  Conference on Robotics and Automation (ICRA)}, pp. 4018--4025, 2022.

\bibitem{nerf_grasp}
J.~Ichnowski*, Y.~Avigal*, J.~Kerr, and K.~Goldberg, ``{Dex-NeRF}: Using a
  neural radiance field to grasp transparent objects,'' in \emph{Conference on
  Robot Learning (CoRL)}, 2020.

\bibitem{nice_slam}
Z.~Zhu, S.~Peng, V.~Larsson, W.~Xu, H.~Bao, Z.~Cui, M.~R. Oswald, and
  M.~Pollefeys, ``Nice-slam: Neural implicit scalable encoding for slam,''
  \emph{2022 IEEE/CVF Conference on Computer Vision and Pattern Recognition
  (CVPR)}, pp. 12\,776--12\,786, 2021.

\bibitem{nerf_navigation}
M.~Adamkiewicz, T.~Chen, A.~Caccavale, R.~Gardner, P.~Culbertson, J.~Bohg, and
  M.~Schwager, ``Vision-only robot navigation in a neural radiance world,''
  \emph{IEEE Robotics and Automation Letters}, vol.~PP, pp. 1--1, 2021.

\bibitem{nerf_supervision}
L.~Yen-Chen, P.~R. Florence, J.~T. Barron, T.-Y. Lin, A.~Rodriguez, and
  P.~Isola, ``Nerf-supervision: Learning dense object descriptors from neural
  radiance fields,'' \emph{2022 International Conference on Robotics and
  Automation (ICRA)}, pp. 6496--6503, 2022.

\bibitem{deep_ensemble}
B.~Lakshminarayanan, A.~Pritzel, and C.~Blundell, ``Simple and scalable
  predictive uncertainty estimation using deep ensembles,'' in \emph{NIPS},
  2017.

\bibitem{mc_dropout}
Y.~Gal and Z.~Ghahramani, ``Dropout as a bayesian approximation: Representing
  model uncertainty in deep learning,'' in \emph{ICML}, 2016.

\bibitem{snerf}
J.~Shen, A.~Ruiz, A.~Agudo, and F.~Moreno-Noguer, ``Stochastic neural radiance
  fields: Quantifying uncertainty in implicit 3d representations,'' in
  \emph{3DV}, 2021.

\bibitem{cfnerf}
J.~Shen, A.~Agudo, F.~Moreno-Noguer, and A.~Ruiz, ``Conditional-flow nerf:
  Accurate 3d modelling with reliable uncertainty quantification,'' in
  \emph{ECCV}, 2022.

\bibitem{niko}
N.~S{\"u}nderhauf, J.~Abou-Chakra, and D.~Miller, ``Density-aware nerf
  ensembles: Quantifying predictive uncertainty in neural radiance fields,''
  \emph{2023 IEEE International Conference on Robotics and Automation (ICRA)},
  2023.

\bibitem{dvgo}
C.~Sun, M.~Sun, and H.-T. Chen, ``Direct voxel grid optimization: Super-fast
  convergence for radiance fields reconstruction,'' in \emph{CVPR}, 2022.

\bibitem{nerf_diet}
A.~Jain, M.~Tancik, and P.~Abbeel, ``Putting nerf on a diet: Semantically
  consistent few-shot view synthesis,'' \emph{2021 IEEE/CVF International
  Conference on Computer Vision (ICCV)}, pp. 5865--5874, 2021.

\bibitem{imap}
E.~Sucar, S.~Liu, J.~Ortiz, and A.~J. Davison, ``imap: Implicit mapping and
  positioning in real-time,'' \emph{2021 IEEE/CVF International Conference on
  Computer Vision (ICCV)}, pp. 6209--6218, 2021.

\bibitem{Magic3D}
C.-H. Lin, J.~Gao, L.~Tang, T.~Takikawa, X.~Zeng, X.~Huang, K.~Kreis,
  S.~Fidler, M.-Y. Liu, and T.-Y. Lin, ``Magic3d: High-resolution text-to-3d
  content creation,'' \emph{2023 IEEE/CVF Conference on Computer Vision and
  Pattern Recognition (CVPR)}, pp. 300--309, 2022.

\bibitem{MeshDiffusion}
Z.~Liu, Y.~Feng, M.~J. Black, D.~Nowrouzezahrai, L.~Paull, and W.~yu~Liu,
  ``Meshdiffusion: Score-based generative 3d mesh modeling,'' \emph{ArXiv},
  vol. abs/2303.08133, 2023.

\bibitem{nerf_audio}
Y.~Guo, K.~Chen, S.~Liang, Y.~Liu, H.~Bao, and J.~Zhang, ``Ad-nerf: Audio
  driven neural radiance fields for talking head synthesis,'' \emph{2021
  IEEE/CVF International Conference on Computer Vision (ICCV)}, pp. 5764--5774,
  2021.

\bibitem{plenoxels}
A.~Yu, S.~Fridovich-Keil, M.~Tancik, Q.~Chen, B.~Recht, and A.~Kanazawa,
  ``Plenoxels: Radiance fields without neural networks,'' in \emph{CVPR}, 2022.

\bibitem{instantNGP}
T.~M{\"u}ller, A.~Evans, C.~Schied, and A.~Keller, ``Instant neural graphics
  primitives with a multiresolution hash encoding,'' \emph{ACM Transactions on
  Graphics (TOG)}, vol.~41, pp. 1 -- 15, 2022.

\bibitem{tensorf}
A.~Chen, Z.~Xu, A.~Geiger, J.~Yu, and H.~Su, ``Tensorf: Tensorial radiance
  fields,'' in \emph{ECCV}, 2022.

\bibitem{unc_review}
M.~Abdar, F.~Pourpanah, S.~Hussain, D.~Rezazadegan, L.~Liu, M.~Ghavamzadeh,
  P.~W. Fieguth, X.~Cao, A.~Khosravi, U.~R. Acharya, V.~Makarenkov, and
  S.~Nahavandi, ``A review of uncertainty quantification in deep learning:
  Techniques, applications and challenges,'' \emph{Inf. Fusion}, vol.~76, pp.
  243--297, 2020.

\bibitem{unc_camera}
A.~Kendall and R.~Cipolla, ``Modelling uncertainty in deep learning for camera
  relocalization,'' \emph{2016 IEEE International Conference on Robotics and
  Automation (ICRA)}, pp. 4762--4769, 2015.

\bibitem{unc_robotics}
A.~Loquercio, M.~Segu, and D.~Scaramuzza, ``A general framework for uncertainty
  estimation in deep learning,'' \emph{IEEE Robotics and Automation Letters},
  vol.~5, pp. 3153--3160, 2019.

\bibitem{bnn_1}
R.~M. Neal, \emph{Bayesian Learning for Neural Networks}.\hskip 1em plus 0.5em
  minus 0.4em\relax Springer New York, NY, 1995.

\bibitem{nerfw}
R.~Martin-Brualla, N.~Radwan, M.~S.~M. Sajjadi, J.~T. Barron, A.~Dosovitskiy,
  and D.~Duckworth, ``{NeRF in the Wild: Neural Radiance Fields for
  Unconstrained Photo Collections},'' in \emph{CVPR}, 2021.

\bibitem{activenerf}
X.~Pan, Z.~Lai, S.~Song, and G.~Huang, ``Activenerf: Learning where to see with
  uncertainty estimation,'' in \emph{European Conference on Computer Vision},
  2022.

\bibitem{UncertaintyGP}
S.~Lee, L.~Chen, J.~Wang, A.~Liniger, S.~Kumar, and F.~Yu, ``Uncertainty guided
  policy for active robotic 3d reconstruction using neural radiance fields,''
  \emph{IEEE Robotics and Automation Letters}, vol.~7, pp. 12\,070--12\,077,
  2022.

\bibitem{ActiveRMAP}
H.~Zhan, J.~Zheng, Y.~Xu, I.~D. Reid, and H.~Rezatofighi, ``Activermap:
  Radiance field for active mapping and planning,'' \emph{ArXiv}, vol.
  abs/2211.12656, 2022.

\bibitem{ActiveIO}
D.~Yan, J.~Liu, F.~Quan, H.~Chen, and M.-Y. Fu, ``Active implicit object
  reconstruction using uncertainty-guided next-best-view optimization,''
  \emph{IEEE Robotics and Automation Letters}, vol.~8, pp. 6395--6402, 2023.

\bibitem{kde}
J.~Jarnicka, ``Multivariate kernel density estimation with a parametric
  support,'' \emph{Opuscula Mathematica}, vol.~29, pp. 41--55, 2009.

\bibitem{mipnerf360}
J.~T. Barron, B.~Mildenhall, D.~Verbin, P.~P. Srinivasan, and P.~Hedman,
  ``Mip-nerf 360: Unbounded anti-aliased neural radiance fields,'' \emph{2022
  IEEE/CVF Conference on Computer Vision and Pattern Recognition (CVPR)}, 2021.

\bibitem{ause}
C.~Qu, W.~Liu, and C.~J. Taylor, ``Bayesian deep basis fitting for depth
  completion with uncertainty,'' in \emph{ICCV}, 2021.

\bibitem{dvgo_v2}
C.~Sun, M.~Sun, and H.-T. Chen, ``Improved direct voxel grid optimization for
  radiance fields reconstruction,'' \emph{ArXiv}, vol. abs/2206.05085, 2022.

\bibitem{nerf++}
K.~Zhang, G.~Riegler, N.~Snavely, and V.~Koltun, ``Nerf++: Analyzing and
  improving neural radiance fields,'' \emph{ArXiv}, vol. abs/2010.07492, 2020.

\end{thebibliography}
}


\end{document}